%% file: main.tex
\documentclass[10pt,twocolumn,letterpaper]{article}
\usepackage[accsupp]{axessibility}
\usepackage{cvpr}              % To produce the CAMERA-READY version
\input{preamble}

\definecolor{cvprblue}{rgb}{0.21,0.49,0.74}
\usepackage[pagebackref,breaklinks,colorlinks,allcolors=cvprblue]{hyperref}

\def\paperID{7952} % *** Enter the Paper ID here
\def\confName{CVPR}
\def\confYear{2026}

\title{Gaussian Mapping for Evolving Scenes}

\author{
Vladimir Yugay$^{1}$\thanks{Equal contrib. Corresp.: \texttt{\{vladimir.yugay, thies.kersten\}@uva.nl}}, 
Thies Kersten$^{1*}$, 
Luca Carlone$^{3}$, 
Theo Gevers$^{1}$, 
Martin R. Oswald$^{1}$, 
Lukas Schmid$^{2,3}$ \\
$^1$University of Amsterdam \quad $^2$University of Technology Nuremberg \quad $^3$Massachusetts Institute of Technology \\
\href{https://vladimiryugay.github.io/game}{\nolinkurl{vladimiryugay.github.io/game}}
}

\begin{document}
\twocolumn[{%
  \vspace{-2em}
  \maketitle
  \captionsetup{type=figure} %suppresses caption warning
  \includegraphics[width=\textwidth]{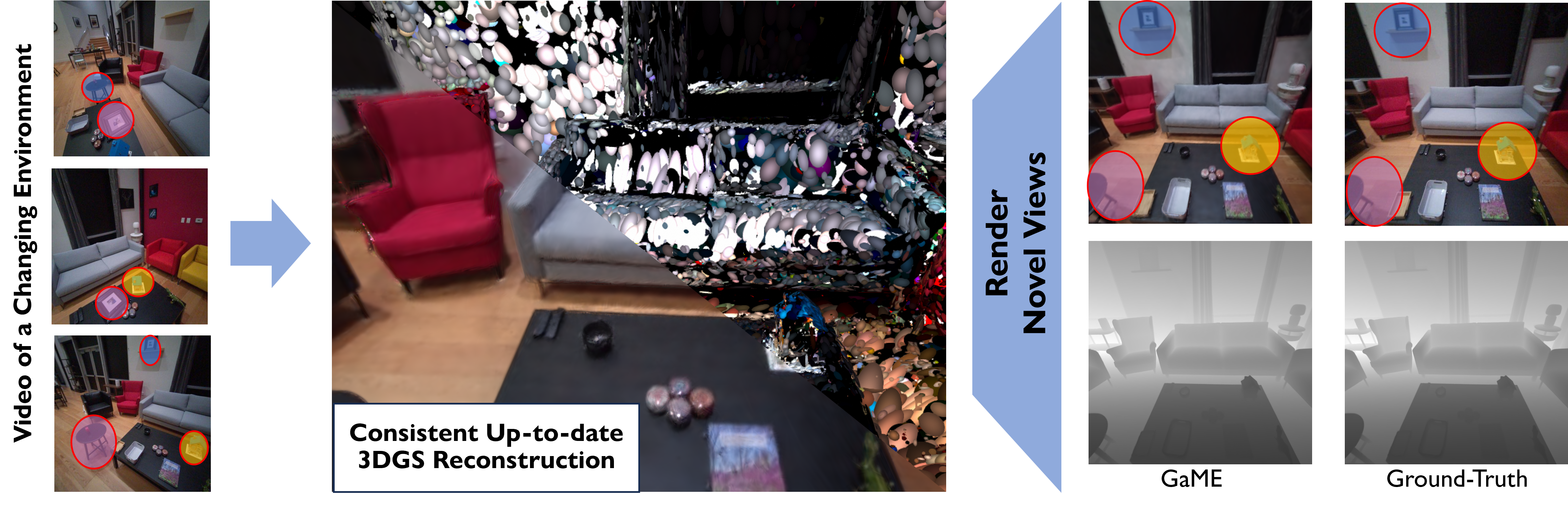}  
  \captionof{figure}{\textbf{\ours} is a dense mapping method capable of novel view synthesis.
  Given a single-camera posed RGBD input stream of an evolving environment, \textit{i.e.}, structural changes happening outside of the camera view but during data capture, \ours reconstructs a consistent up-to-date 3D Gaussian map that can be rendered from previously unseen viewpoints. 
  We showcase the high-fidelity 3D Gaussian map of a real-world environment, where our method effectively detects changes and accurately integrates them into the map. 
  Our model accurately renders fine-grained objects that were moved during the recording, highlighted in blue, yellow, and purple.
  }
  \label{fig:teaser}
  \vspace{1em}
}]
\input{sec/0_abstract}    
\input{sec/1_intro}
\input{sec/2_related_work}
\input{sec/3_3dgs}
\input{sec/4_method}
\input{sec/5_experiments}
\input{sec/6_limitations}
\input{sec/7_conclusion}

{
    \small
    \bibliographystyle{ieeenat_fullname}
    \bibliography{main}
}

% WARNING: do not forget to delete the supplementary pages from your submission 
% \input{sec/X_suppl}

\end{document}

%% file: preamble.tex
\usepackage[table]{xcolor} % ensure table coloring is enabled
\colorlet{colorFst}{Green!25}       % first
\colorlet{colorSnd}{SpringGreen!45} % second
\colorlet{colorTrd}{Yellow!30}      % third
\colorlet{colorLow}{darkgray!30}    % low-light color

\newcommand{\fs}{\cellcolor{colorFst}\bf}   % first
\newcommand{\nd}{\cellcolor{colorSnd}}      % second
\newcommand{\rd}{\cellcolor{colorTrd}}      % third

\newcommand{\fsttxt}[1]{\colorbox{colorFst}{\textbf{#1}}}
\newcommand{\sndtxt}[1]{\colorbox{colorSnd}{#1}}
\newcommand{\trdtxt}[1]{\colorbox{colorTrd}{#1}}

\newcommand{\ours}{GaME\xspace}
\newcommand{\boldparagraph}[1]{\vspace{2pt}\noindent{\bf #1}}

%% file: sec/0_abstract.tex
\begin{abstract}
Mapping systems with novel view synthesis (NVS) capabilities, most notably 3D Gaussian Splatting (3DGS), are widely used in computer vision, as well as in various applications, including augmented reality, robotics, and autonomous driving. 
However, many current approaches are limited to static scenes. 
While recent works have begun addressing short-term dynamics (motion within the camera's view), long-term dynamics (the scene evolving through changes out of view) remain less explored.
To overcome this limitation, we introduce a dynamic scene adaptation mechanism to continuously update 3DGS to reflect the latest changes. 
Since maintaining consistency remains challenging due to stale observations disrupting the reconstruction process, we further propose a novel keyframe management mechanism that discards outdated observations while preserving as much information as possible. We thoroughly evaluate Gaussian Mapping for Evolving Scenes (\ours) on both synthetic and real-world datasets, achieving a 29.7\% improvement in PSNR and a $3\times$-improvement in L1 depth error over the most competitive baseline.
\end{abstract}

%% file: sec/1_intro.tex
\section{Introduction}

Visual mapping reconstructs a consistent 3D representation of an environment from monocular or multi-view imagery. 
It is foundational for perception and decision-making in autonomous driving, mobile robotics, and AR/VR, where spatial understanding is required for navigation, planning, and interaction. 
Recent advances in visual dense reconstruction~\cite{rosinol2021kimera,murai2024_mast3rslam,schmid2024rss_khronos}, have improved robustness and accuracy in complex, real-world scenes. 
Nevertheless, significant challenges persist in handling changing environments.

Recent mapping systems have been enhanced with novel view synthesis (NVS) capabilities~\cite{murai2024monogs,keetha2024splatam}, enabling them to render realistic views of scenes, allowing for detailed scene exploration and the creation of high-quality virtual environments. 
These approaches typically assume that the scene is \emph{static}, such that optimization over multiple frames is well-conditioned.
However, real-world environments are predominantly dynamic and include both \emph{short-term} dynamic effects (\ie, things moving \emph{within} view of the camera) as well as \emph{long-term} dynamics (\ie, the scene evolving through changes \emph{outside} the view of the camera; for a detailed definition see \cite{sh-ch15-dyndef}).
While some of the most recent NVS approaches address short-term dynamic objects~\cite{zheng2025wildgs, xu2024dgslam}, long-term dynamics remain less explored. 
As a result, modern reconstruction methods with NVS capabilities struggle to capture changes as the scene evolves during data capture but outside the camera view, resulting in corrupted and erroneous reconstruction.

During dense mapping of \emph{long-term dynamic} scenes, systems must continuously update the 3D representation to capture the scene's evolution and maintain reliable operation. 
This is especially challenging compared to classical \emph{multi-session} mapping, as changes can occur \emph{at any time} during data collection.
Recently, initial methods have started to address this challenge. 
A first approach is presented in Panoptic Multi-TSDFs~\cite{schmid2022panoptic}, which builds semantically consistent submaps and reasons about changes at the submap level.
The following works~\cite{fu2022robust,qian2022pocd,schmid2024rss_khronos} similarly focus on object-level mapping to handle evolving scenes.
However, all these methods rely on map representations that cannot easily be re-rendered, preventing their use in AR/VR or digital maps where realistic rendering is essential.

This work addresses the challenge of NVS mapping using 3DGS in evolving, long-term dynamic scenes. The key insight of our method is that two main problems prevent successful 3DGS mapping in changing environments: an outdated map and stale observations. We address these limitations by introducing a dynamic scene adaptation (DSA) mechanism that detects changes and integrates them into the 3DGS map, as well as a keyframe management system that removes stale observations while minimizing the loss of information.
% The key insight of our approach is that environmental changes are not random but typically follow semantically consistent patterns.
% We therefore integrate semantic consistency with the inherent properties of 3DGS to efficiently detect and adapt the incrementally built 3DGS model. As multi-view optimization is essential for accurate 3DGS mapping, we introduce a keyframe management method that appropriately masks stale areas in keyframes to retain useful information while accounting for changes, resulting in a well-conditioned 3DGS optimization process in evolving scenes.
We make the following contributions:
\begin{itemize}
    \item We present \ours, the first NVS-capable mapping system addressing evolving, long-term dynamic scenes.
    \item A Dynamic Scene Adaptation (DSA) mechanism to incrementally update a 3DGS model.
    \item An efficient keyframe management strategy for accurate 3D reconstruction through long-term scene changes.
    \item We thoroughly evaluate \ours on synthetic and real-world data, showing a performance increase of 325\% in depth and 29\% in color rendering over the most competitive baselines. We release the code open-source.
\end{itemize}
 

%% file: sec/2_related_work.tex
\section{Related Work}

\boldparagraph{3D Change Detection.}
The goal of change detection is to handle long-term dynamic effects, \ie, changes to the scene occurring \emph{outside} the view of the sensor.
Typically, this is addressed in a \emph{multi-session} setting.
Once a map for each session is constructed, changes can be identified through \emph{scene differencing}~\cite{ambrucs2014iros_meta,fehr2017icra_tsdfchange,kim2022lt,rowell2024lista}. 
To achieve an object-level change understanding, this has been extended using semantic information~\cite{langer2020robust}, where recent trends focus on the extraction of more specialized object features, including learned shape descriptors~\cite{wald2019rio, fu2022planesdf, rowell2024lista}, neural object representations~\cite{fu2022robust, zhu2024living}, and language embeddings~\cite{rashid2024lifelong}.
Most related to us, \citet{lu20253dgs} recently presented a 3D Gaussian Splatting-based~\cite{kerbl2023gaussian_splatting} approach by re-rendering the scene to newly collected views and using EfficientSAM~\cite{xiong2024efficientsam} for 2D change detection. 
However, the assumption that the scene does not change during each session and offline processing is highly limiting, as changes in evolving scenes can occur \emph{at any time} during the mapping process. 
% yes, sure BUT it is implied but not directly mentioned why this is important. e.g. It thus has more practical applications / Most real world data is rarely static. Essentially we could sell even more why one should care.
In contrast, our approach is designed to operate online and does not impose any limitations on when the scene changes.

\boldparagraph{Online Reconstruction of Evolving Scenes.} 
Recently, methods addressing the modeling of evolving scenes in an online fashion have emerged. A first approach is presented by \citet{schmid2022panoptic}, which generates local semantically consistent submaps and incrementally reasons about changes at the submap level.
\citet{fu2022robust} proposes neural object descriptors to build an object-level pose graph and detect changes in the graph configuration.
This has recently been extended with $SE(3)$-equivariant descriptors~\cite{fu2023neuse}.
\citet{qian2022pocd} present a frame-to-map-tracking approach, using a probabilistic update rule to detect object-level changes, which is further extended to a variational factor-graph approach in~\cite{qian2023pov}.
A unified short- and long-term dynamic reconstruction formulation is presented in Khronos~\cite{schmid2024rss_khronos}, where objects are reconstructed locally and changes detected using a library of rays. 
Nonetheless, these methods rely on map representations that do not support realistic rendering. 
In contrast, \ours provides scene reconstruction capable of real-time color and depth rendering.

\boldparagraph{Dynamic Gaussian Splatting.}
3D Gaussian Splatting ~\cite{kerbl2023gaussian_splatting} has revolutionized novel view synthesis (NVS) by enabling photorealistic, real-time rendering at over 100 FPS. 
Compared to neural radiance fields~\cite{mildenhall2021nerf}, 3DGS is significantly more memory-efficient and faster to optimize. 
The problem of dynamic or 4D GS has attracted broad interest.
A series of works~\cite{das2024neural,luiten2023dynamic,wu20244dgs,yang2023deformable3dgs} optimize a canonical set of Gaussians from the initial frame and model temporal variations through a deformation field. However, these methods are limited to short video sequences, as they cannot introduce new Gaussians after the initial frame. 
Another class of approaches~\cite{duan_2024_4drotorgs,kitsumata2024compact3dgs,yang2023gs4d} directly models temporal Gaussians that can exist over subsets of frames. Despite their improved flexibility, these methods still require offline optimization and multi-view input, which makes scalability a significant bottleneck for high-quality dynamic reconstruction. 
In contrast, our mapping operates online using only a single RGB-D camera.

\boldparagraph{Online Gaussian Mapping.}
Most related to us, 3D Gaussian Splatting has sparked a wave of online RGB-D mapping methods~\cite{murai2024monogs,keetha2024splatam,yan2023gs,hhuang2024photoslam,yugay2023gaussianslam} that adopt 3D Gaussians as their primary scene representation. 
While these methods perform well in static scenes, they struggle in dynamic environments. 
More recent approaches~\cite{xu2024dgslam,zheng2025wildgs} tackle \textit{short-term} dynamics within the camera’s view by incorporating segmentation models to suppress transient objects, such as humans or small moving elements. 
However, they remain limited in addressing \textit{long-term} scene changes, as stale observations corrupt the optimization process. 
In contrast, \ours is designed to robustly handle evolving environments by filtering outdated information and maintaining high rendering quality throughout reconstruction.

%% file: sec/3_3dgs.tex
\section{Background: 3D Gaussian Splatting}

3D Gaussian splatting (3DGS)~\citep{kerbl2023gaussian_splatting} is an effective method for representing 3D scenes with novel-view synthesis capability.
This approach is notable for its speed, without compromising the rendering quality. In~\citet{kerbl2023gaussian_splatting},
3D Gaussians are initialized from sparse Structure-from-Motion points % point cloud of a scene.
%With images observing the scene from different angles, the Gaussian parameters 
and are further optimized using differentiable rendering from multiple views.
% During training, 3D Gaussians are adaptively added or removed to better render the images based on a set of heuristics.
Each Gaussian is parameterized by mean $\mu \in \mathbb{R}^3$,
covariance $\Sigma \in \mathbb{R}^{3 \times 3}$, opacity $o \in \mathbb{R}$,
and RGB color $C \in \mathbb{R}^3$. The mean of a projected (splatted) 3D Gaussian in the 2D
image plane $\mu^{I}$ is computed as follows:
\begin{equation}
    \mu^{I} = \pi\big(P(T_{wc} \mu_\text{homogeneous} )\big) \enspace,
\end{equation}
where $T_{wc} \in SE(3)$ is the world-to-camera transformation, $P \in \mathbb{R}^{4 \times 4}$ is an OpenGL-style projection matrix, % OpenGL-style
$\pi: \mathbb{R}^{4} \rightarrow \mathbb{R}^{2}$ is a projection to pixel coordinates. 
The 2D covariance $\Sigma^{I}$ of a splatted Gaussian is:
\begin{equation}
    \Sigma^{I} = J R_{wc} \Sigma R_{wc}^T J^T \enspace,
\end{equation}
where $J \in \mathbb{R}^{2 \times 3}$ is an affine transformation from \citet{zwicker2001surface}, $R_{wc} \in SO(3)$ is
the rotation component of world-to-camera transformation $T_{wc}$. We refer to \citet{zwicker2001surface} for further details
about the projection matrices. The color $C$ along one channel $ch$ at a pixel $i$ is influenced by $m$ ordered Gaussians an rendered as:
\begin{align}
C^{ch}_{i}
&= \sum_{j \le m} C^{ch}_{j}\,\alpha_{j}
   \prod_{k < j} (1 - \alpha_{k}), \nonumber\\
\text{where}\quad
\alpha_{j} &= o_{j}\,e^{-\sigma_{j}}, \qquad
\sigma_{j} = \tfrac{1}{2}\,\Delta_{j}^\top
\Sigma_{j}^{-1}\Delta_{j}.
\label{eq:color_blending}
\end{align}

where $\Delta_{j} \in \mathbb{R}^{2}$ is the offset between the pixel coordinates and the 2D mean of a splatted Gaussian. 
The parameters of the 3D Gaussians are iteratively optimized by minimizing the photometric loss between rendered and training images.
During optimization, $C$ is encoded with spherical harmonics $SH \in \mathbb{R}^{15}$ to account for direction-based color variations.
Covariance is decomposed as $\Sigma = RSS^{T}R^{T}$, where $R \in SE(3)$
and $S = \text{diag}(s) \in \mathbb{R}^{3 \times 3}$ are rotation and scale, respectively, to preserve the covariance positive semi-definite property during gradient-based optimization.

%% file: sec/4_method.tex
\section{Method}

\begin{figure*}[ht!]
  \centering
  \includegraphics[width=\linewidth]{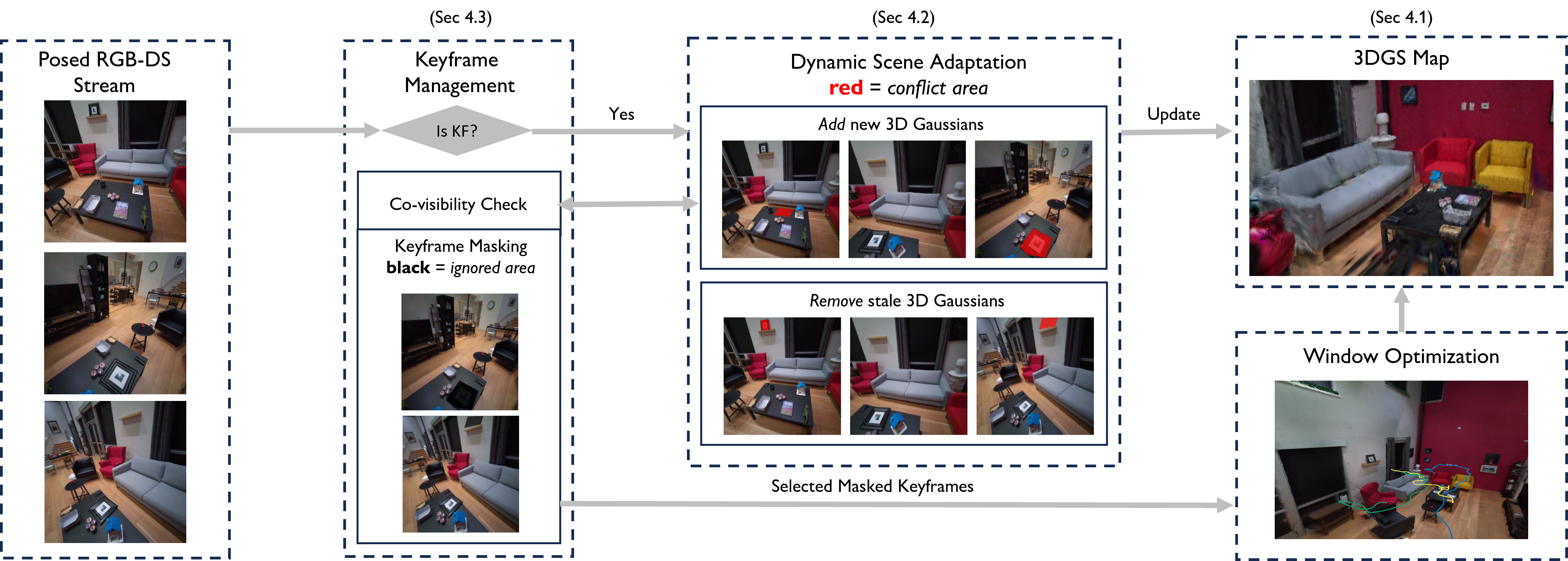}
  \caption{\textbf{\ours Architecture.} Given a segmented RGB-D input stream, the keyframe management system selects keyframes and triggers the dynamic scene adaptation (DSA) module. DSA first integrates newly observed geometry, then removes outdated geometry using covisible keyframes from the 3D Gaussian Splatting map. The keyframe manager then masks stale regions, and the mapping system uses the processed keyframes for local covisibility window optimization.}
  \label{fig:architecture}
\end{figure*}

%
%------------------------------------------------------------------------------
\ours builds and maintains a 3D map capable of novel view synthesis of a static environment undergoing structural changes happening outside the sensor's view at any time. Our method processes posed depth and color images from a single monocular RGB-D sensor. For each keyframe, \ours triggers the Dynamic Scene Adaptation (DSA) module to incorporate changed, removed, or added geometry in the global map. Simultaneously, the keyframe management system updates the observations to ensure they reflect the scene's up-to-date state and do not disrupt the optimization process. An overview of our approach is shown in~\cref{fig:architecture}.
\subsection{Online Mapping}
\label{sec:online_mapping}
The primary goal of our method is to create a 3DGS map that is consistent with the most recent observations. Unlike previous work~\cite{schmid2022panoptic} operating on the point clouds, we chose not to model objects as separate entities, as this is prohibitively expensive in a 3DGS setup. Moreover, accurately optimizing individual objects represented as 3DGS is hard due to the nature of the splatting mechanism \eqref{eq:color_blending}, which inherently correlates all Gaussians. Instead, we propose to build a singular representation consisting of a set of 3D Gaussians $\{\mathbf{G}_i\}_{i=1}^{N}$ and extract changed geometry on an `as needed' basis during \textit{Dynamic Scene Adaptation} (\cref{sec:dynamic_scene_adaptation}).
During online mapping, we optimize the 3DGS parameters of the scene using a set of keyframes for supervision, minimizing the loss:
\begin{align}
    L = L_\text{color}(\hat{I}, I) + L_\text{depth}(\hat{D}, D),
\label{eq:joint_loss}
\end{align}
where $I$ is the original image, $\hat{I}$ is the rendered image, $D$ and $\hat{D}$ are the measured and reconstructed depth maps.  The color loss and depth losses are defined as:
\begin{align}
L_\text{color}(\hat{I}, I)
&= (1 - \lambda)\,
   \frac{1}{K} \sum_{p}
   \big| \hat{I}(p) - I(p) \big| \nonumber\\
&\quad+\, \lambda\,\big(1 - \mathrm{SSIM}(\hat{I}, I)\big),
\label{eq:color_loss}
\end{align}

\begin{align}
L_\text{depth}(\hat{D}, D) = \frac{1}{K} \sum_{p} |\hat{D}(p) - D(p)|,
\label{eq:depth_loss}
\end{align}
where $K$ is the number of rendered pixels in the rendered image or depth map, $p \in \mathbb{Z}^2$ denotes the pixel coordinates, and $\mathrm{SSIM}$ is a structure similarity loss~\cite{wang2004image}, and $\lambda \in [0, 1]$ is the weight balancing the L1 and SSIM terms. We additionally include an isotropic regularization term~\cite{yugay2023gaussianslam} to prevent degenerate Gaussian shapes.

%%%%%%%%%%%%%%%%%%%%%%%%%%%%%%%%%%%%%%%%%%%%%%%%%%%%%%%%%%
\subsection{Dynamic Scene Adaptation}
\label{sec:dynamic_scene_adaptation}
In addition to reconstructing newly explored regions of the scenes, \ours addresses three potential scenarios: the addition, movement, or removal of rigid geometry outside of the camera view. 
While \ours can be readily combined with object re-localization techniques~\cite{wald2019rio, fu2023neuse, rowell2024lista}, object tracking does not affect NVS, and the problem can thus be simplified into two core operations: \emph{Add} and \emph{Remove}. 

\boldparagraph{Add}. 
As the first step, DSA utilizes the current observations to detect changes in the scene where new geometry has been added. They are defined as already reconstructed Gaussians where the observed depth is closer to the camera than the rendered depth from the model:
\begin{align}
\mathcal{G}_\text{add}
&= \Big\{\, \mathbf{G}_i \,\Big|\, 
   \alpha(\mathbf{p}) \geq \epsilon_\text{opacity} \,\wedge \nonumber\\
&\qquad
   \hat{D}(\mathbf{p}) > D(\mathbf{p}) + \epsilon_{\text{depth}}
   \,\Big\},
\label{eq:add}
\end{align}
where $\alpha(\mathbf{p})$ represents rendered opacity at pixel $\mathbf{p} \in \mathbb{Z}^2$,  $\epsilon_\text{opacity}, \epsilon_{\text{depth}} \in \mathbb{R}$ are the thresholds accounting for the rendering noise.

On the next step, DSA dynamically seeds new Gaussian primitives in newly explored regions to reconstruct the complete scene. Following~\cite{yugay2023gaussianslam}, such regions $\mathcal{R}_\text{new}$ are characterized by low opacity, a large rendered depth overshoot, or high color error:
\begin{align}
\mathcal{R}_\text{new}
&= \Big\{\, \mathbf{p} \in \mathbb{Z}^2 \,\Big|\, 
   \alpha(\mathbf{p}) < \epsilon_\text{opacity} \,\vee \nonumber\\
&\qquad
   \hat{D}(\mathbf{p}) - D(\mathbf{p}) > \tau \cdot \mathrm{median}(L_\text{depth}) \,\vee \nonumber\\
&\qquad
   L_\text{color}(\mathbf{p}) > \epsilon_\text{seed}
   \,\Big\}.
\label{eq:new}
\end{align}
where $L_\text{depth}(\mathbf{p}) = |\hat{D}(\mathbf{p}) - D(\mathbf{p})|$ and $L_\text{color}(\mathbf{p}) = |\hat{I}(\mathbf{p}) - I(\mathbf{p})|$ denote the per-pixel depth and color errors, and $\tau \in \mathbb{R}_{>0}$, $\epsilon_\text{seed} \in \mathbb{R}$ are thresholds. The method lifts the input RGB-D frame in $\mathcal{R}_\text{new}$ into 3D space and uses it as the means for new Gaussians. After that, the new Gaussians are optimized against the current keyframe for a small number of iterations (see~\cref{sec:online_mapping}) before the \textit{Remove} step.

\begin{figure*}[t!]
  \centering
  \makebox[\linewidth][c]{\includegraphics[width=\textwidth]{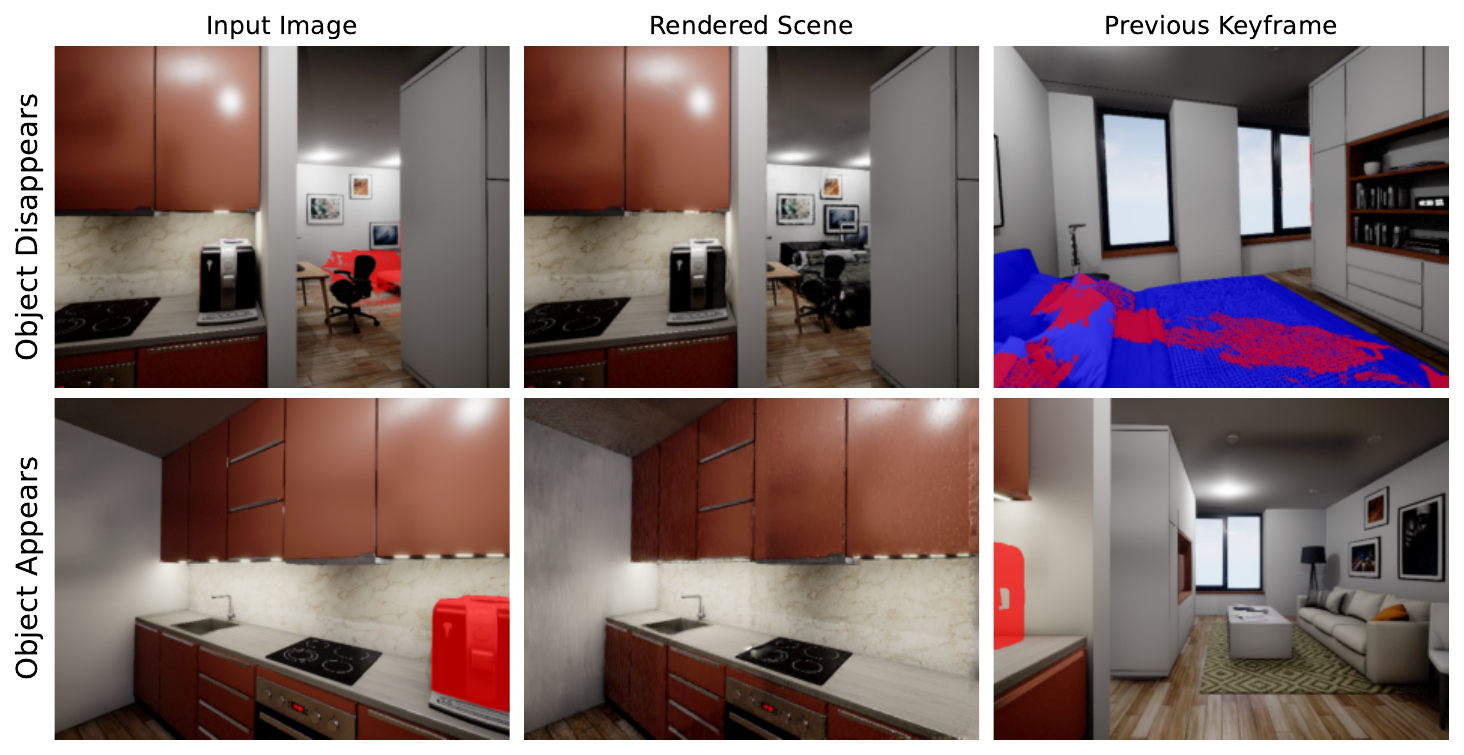}}
  \caption{\textbf{Illustration of \textit{Add} and \textit{Remove} operations.} The input disagrees with the rendered model (red). For disappearance (top), the conflicting region is projected to previous keyframes for removal (red), where complete object consistency is enforced through the object mask (blue). This allows \ours to extract complete objects even under partial observations and occlusion. When a new object appears on the scene (bottom), new Gaussians are added (red) and the area of the new object is marked as stale in previous keyframes to prevent the contamination of the optimization process.}
  \label{fig:conflict_illustration}
\end{figure*}
\boldparagraph{Remove}. On the next step, DSA detects stale Gaussians that need to be removed from the scene representation. 
Specifically, changed parts of the 3DGS model can be identified as those with high opacity, but disagreeing visually or geometrically with the new observations:
\begin{equation}
\begin{aligned}
\mathcal{G}_\text{remove}
&= \Big\{\, \mathbf{G}_i \,\Big|\, 
   \alpha(\mathbf{p}) \geq \epsilon_\text{opacity} \,\wedge \\
&\quad L_\text{color}(\mathbf{p}) > \epsilon_\text{color} \,\wedge \,
   \hat{D}(\mathbf{p}) < D(\mathbf{p}) - \epsilon_\text{depth}
   \,\Big\}
\end{aligned}
\label{eq:gremove}
\end{equation}
where $\epsilon_\text{color} \in \mathbb{R}$ accounts for color rendering noise. There are two notable changes compared to the \textit{Add} condition. 
First, the sign of the depth criterion is inverted, reflecting areas where the rays of the observation would penetrate into the current model, thus indicating that the model geometry can no longer be present. 
This also avoids spurious detections where objects are simply occluded rather than absent.
Second, we leverage the high visual fidelity of 3DGS as another conflict signal by adding a color term.
In contrast to most long-term mapping methods, which rely solely on 3D information~\cite{fehr2017icra_tsdfchange, rowell2024lista, kim2022lt, schmid2022panoptic, schmid2024rss_khronos, qian2023pov}, this allows \ours to also detect changes that are not geometrically significant, such as a picture removed from the wall.

Naively removing the $\mathcal{G}_\text{remove}$ breaks the consistency of the scene. Specifically, when only parts of the objects that need to be removed are seen in the current frame. 
To address this, DSA renders $\mathcal{G}_\text{remove}$ to each covisible keyframe $KF$ and verifies whether the color and depth are consistent with the model:
\begin{align}
\mathcal{R}_\text{remove}^\text{KF} 
&= \Big\{\, \mathbf{p} \in \mathbb{Z}^2 \,\Big|\,
   L_\text{color}(\mathbf{p}) < \epsilon_\text{color}, \nonumber\\
&\qquad
   L_\text{depth}(\mathbf{p}) < \epsilon_\text{depth},\;
   \alpha(\mathbf{p}) \geq \epsilon_\text{opacity}
   \,\Big\}.
\label{eq:rrm_kf}
\end{align}
$\mathcal{R}_\text{remove}^\text{KF}$
highlights the areas of previous keyframes where they observe the Gaussians that are removed. However, the detected conflicting areas might not accurately cover the objects due to occlusions and sensing noise.
To address this, DSA uses dense SAM~\cite{kirillov2023segany} masks of the keyframes, and selects all masks that are intersecting with the rendering of $\mathcal{G}_\text{remove}$. Then the selected masks are used to segment the Gaussians that additionally need to be removed $\mathcal{G}_\text{remove}^\text{KF}$. 
Our method uses FlashSplat~\cite{shen2024flashsplat} to optimally assign masks to the Gaussians, providing multiview consistency making our method robust to the noisy depth and color renderings. 
Notably, SAM masks may overlap and carry no semantic labels; their sole purpose is to delineate the complete silhouette of object that needs to be removed. 
This allows DSA to remove whole objects, even when only parts of them were visible in a current frame.

Finally, the joint set, $\mathcal{G}_\text{remove} \cup \bigcup_\text{KF}\mathcal{G}_\text{remove}^\text{KF}$, is removed from the global Gaussian model, and all the Gaussians are optimized for a small number of iterations to integrate the changes. 
The DSA process is illustrated in \cref{fig:conflict_illustration}.
%%%%%%%%%%%%%%%%%%%%%%%%%%%%%%%%%%%%%

\subsection{Keyframe Management}
\label{subsection_keyframe_mngmt}

Since jointly optimizing Gaussians using all frames from a video stream is computationally infeasible, we maintain a smaller set of keyframes $W_k$. 
Effective keyframe management (KM) aims to select non-redundant keyframes that observe the same region while spanning a wide baseline to enforce stronger multiview constraints. 
Following~\cite{keetha2024splatam,yugay2023gaussianslam}, keyframes are selected every time a frame exceeds a translation $\theta_\text{translation} \in \mathbb{R}$ or a rotation $\theta_\text{rotation} \in \mathbb{R}$ threshold.

\boldparagraph{Covisibility.} 
Upon triggering \textit{Add} or \textit{Remove} operations by DSA, all covisible keyframes are retrieved by reprojecting 3D points from the current frame and selecting frames where these points are not occluded.
This approach differs from conventional 3DGS methods~\cite{murai2024monogs,xu2024dgslam}, which determine covisibility based on the number of Gaussians used to re-render the keyframes. We found projection to be more reliable than re-rendering in multi-room environments, as Gaussians from different rooms may appear visible due to alpha blending, leading to erroneous covisibility estimates.

\boldparagraph{Keyframe Masking}. 
In evolving scenes, excluding outdated visual information plays a critical role, as the scene may change over time, making previous observations detrimental for the 3DGS optimization process.
However, discarding entire keyframes that observe stale geometry can result in losing useful information. While some parts of the scene may change, other regions often remain stable and can provide valuable signals for 3DGS optimization. Therefore, rather than discarding entire keyframes as stale, \ours selectively ignores only the stale areas within them.

\begin{figure}[h]
\centering
\begin{tabular}{cc}
\begin{minipage}[b]{0.45\linewidth}
  \centering
  \includegraphics[width=\linewidth]{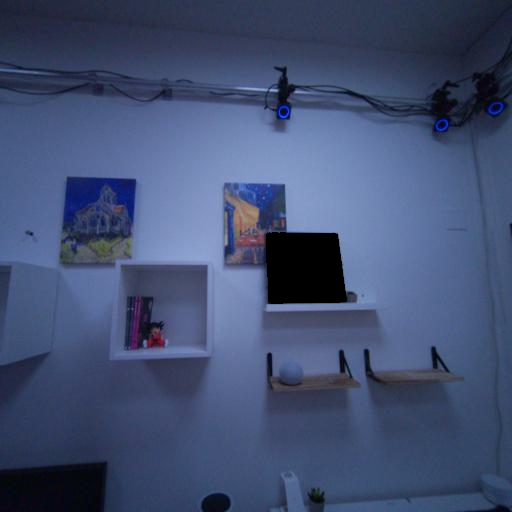}
\end{minipage} &
\begin{minipage}[b]{0.45\linewidth}
  \centering
  \includegraphics[width=\linewidth]{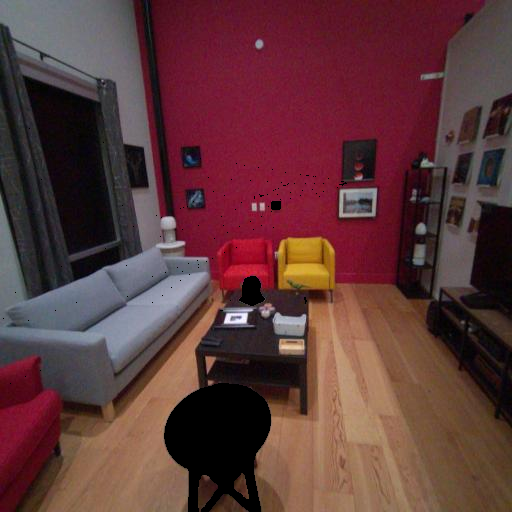}
\end{minipage} \\
\end{tabular}
\caption{\textbf{Visualizations of the outdated keyframes.} After detecting the changes, our keyframe management masks out the areas of covisible keyframes observing changed geometry.
}
\label{fig:masked_out_areas}
\end{figure}

To achieve this, before seeding new Gaussians in $\mathcal{R}_\text{new}$ or removing $\mathcal{G}_\text{remove}$ and $\mathcal{G}_\text{remove}^\text{KF}$, our keyframe management system renders them onto covisible keyframes and assesses conflicts via photometric and geometric losses~\cref{eq:joint_loss}. 
High-error regions correspond to areas of the map that were changed but are not reflected in the observations. To mitigate rendering noise~\cite{kerbl2023gaussian_splatting}, these regions are refined into object-aligned areas before being marked as ignored: for \textit{Remove}, using precomputed SAM~\cite{kirillov2023segany} masks that intersect the conflict region; for \textit{Add}, via morphological closing of the reprojected depth conflict. During optimization, all keyframes with any SAM mask that is sufficiently covered by the erroneous area are ignored. 
For the rest, \ours optimizes the losses from ~\cref{eq:joint_loss} using the masked version of losses. We visualize the keyframes with masked out stale areas on~\cref{fig:masked_out_areas}.
\vspace{-2em}

%% file: sec/5_experiments.tex
\begin{figure*}[ht!]
  \centering
  \newcommand{\wratio}{0.19}
  \newcommand{\imgheight}{2.8cm}

  \makebox[0.01\textwidth][c]{}
  \makebox[\wratio\textwidth]{\normalsize SplaTAM~\cite{keetha2024splatam}}
  \makebox[\wratio\textwidth]{\normalsize MonoGS~\cite{murai2024monogs}}
  \makebox[\wratio\textwidth]{\normalsize DG-SLAM~\cite{xu2024dgslam}}
  \makebox[\wratio\textwidth]{\normalsize \ours (Ours)}
  \makebox[\wratio\textwidth]{\normalsize Ground Truth}
  \\
  \raisebox{1.2cm}{\makebox[0.01\textwidth]{\rotatebox{90}{\normalsize A}}}
  \includegraphics[width=\wratio\textwidth,height=\imgheight]{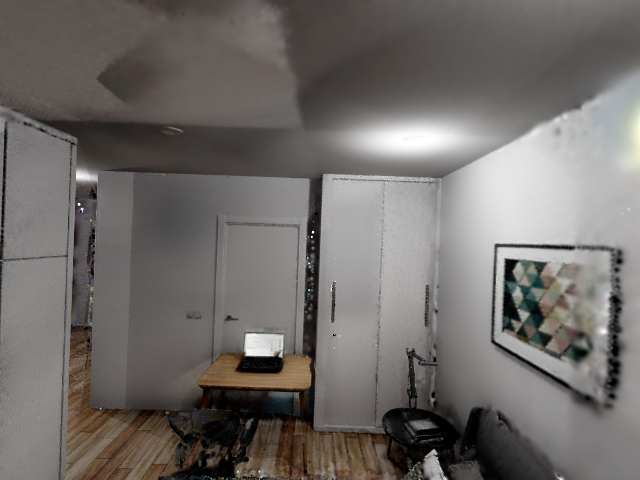}
  \includegraphics[width=\wratio\textwidth,height=\imgheight]{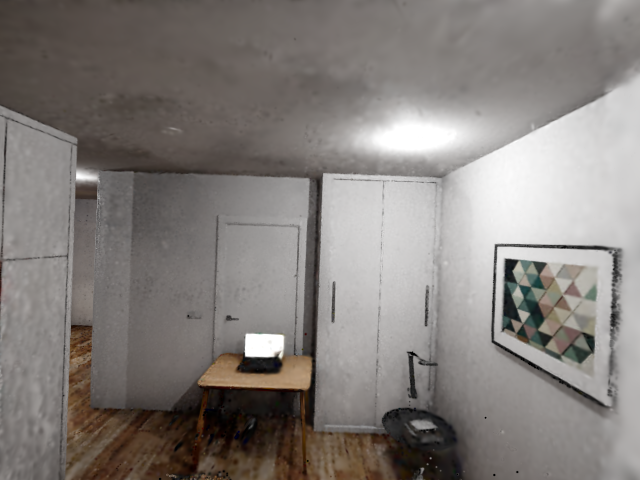}
  \includegraphics[width=\wratio\textwidth,height=\imgheight]{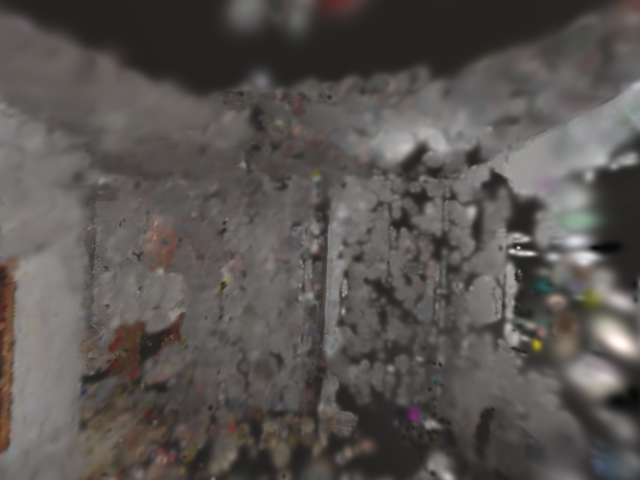}  
  \includegraphics[width=\wratio\textwidth,height=\imgheight]{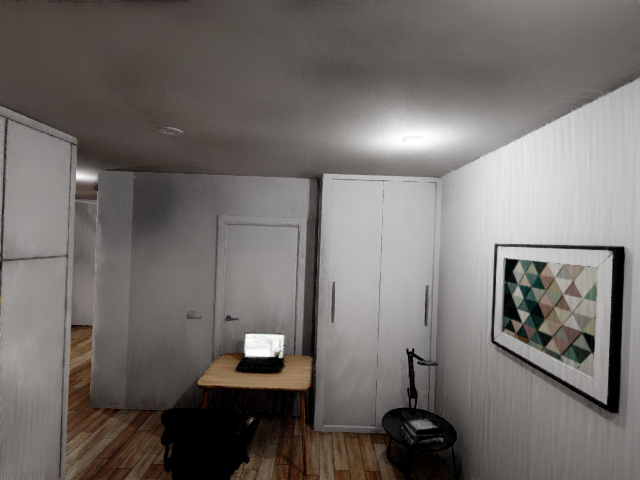}
  \includegraphics[width=\wratio\textwidth,height=\imgheight]{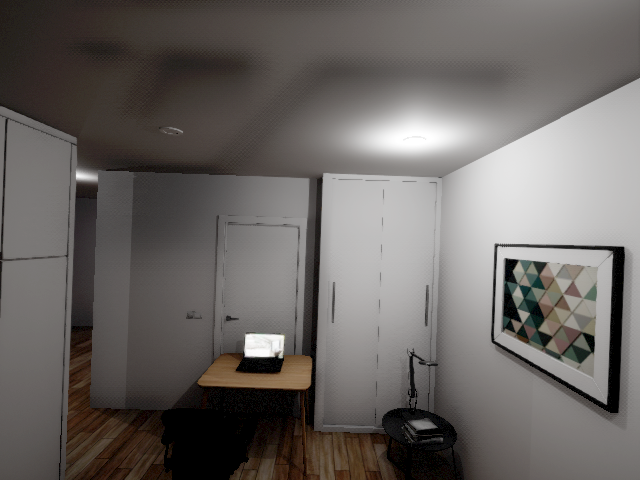}
  \\
  \raisebox{1.2cm}{\makebox[0.01\textwidth]{\rotatebox{90}{\normalsize B}}}
  \includegraphics[width=\wratio\textwidth,height=\imgheight]{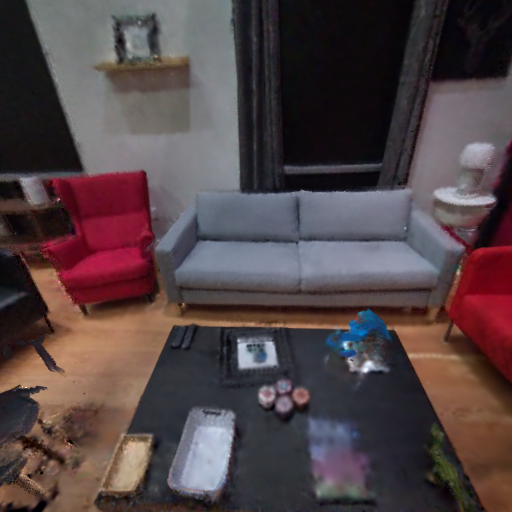}
  \includegraphics[width=\wratio\textwidth,height=\imgheight]{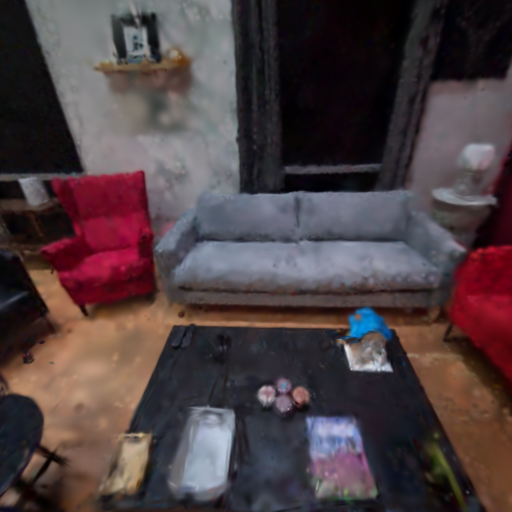}
  \includegraphics[width=\wratio\textwidth,height=\imgheight]{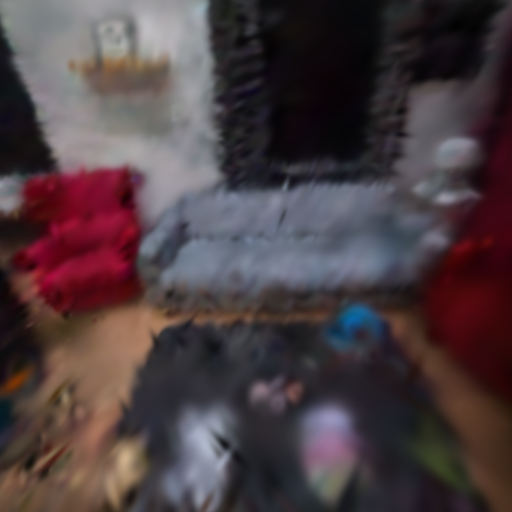}
  \includegraphics[width=\wratio\textwidth,height=\imgheight]{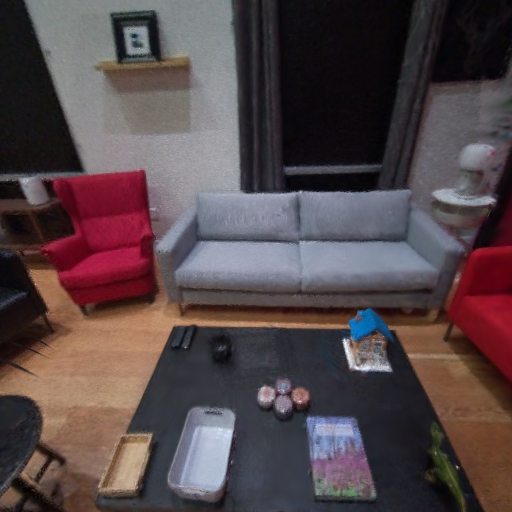}
  \includegraphics[width=\wratio\textwidth,height=\imgheight]{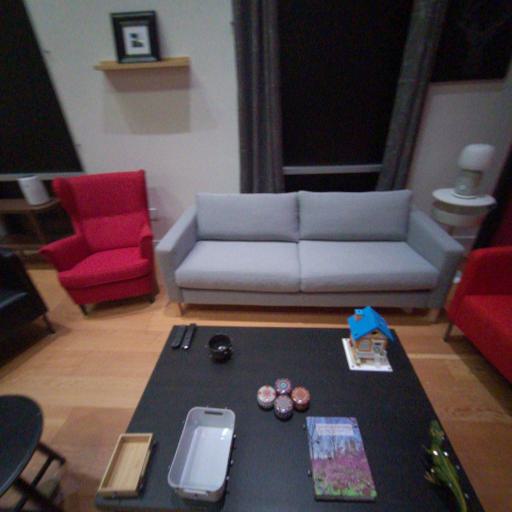}
  \\
  \raisebox{1.2cm}{\makebox[0.01\textwidth]{\rotatebox{90}{\normalsize C}}}
  \includegraphics[width=\wratio\textwidth,height=\imgheight]{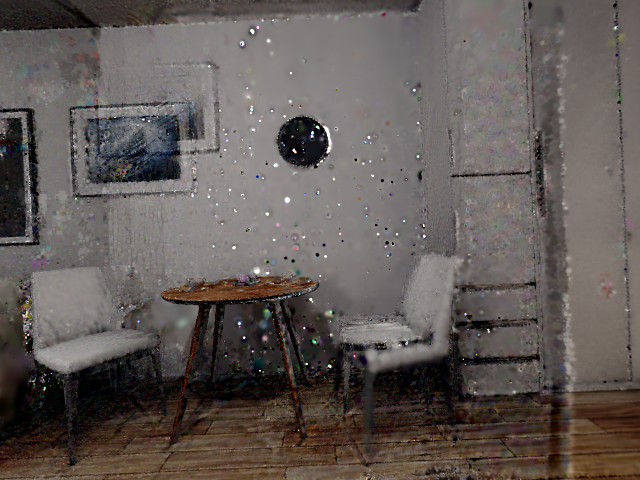}
  \includegraphics[width=\wratio\textwidth,height=\imgheight]{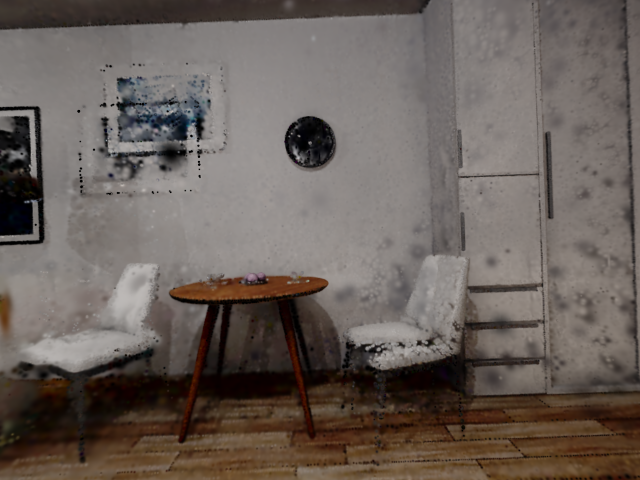}
  \includegraphics[width=\wratio\textwidth,height=\imgheight]{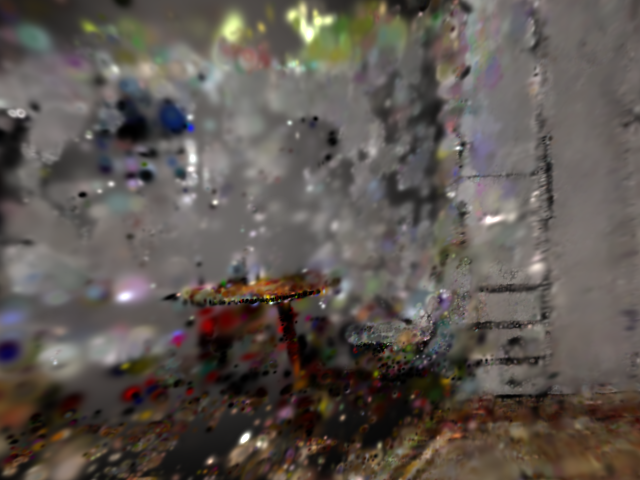}
  \includegraphics[width=\wratio\textwidth,height=\imgheight]{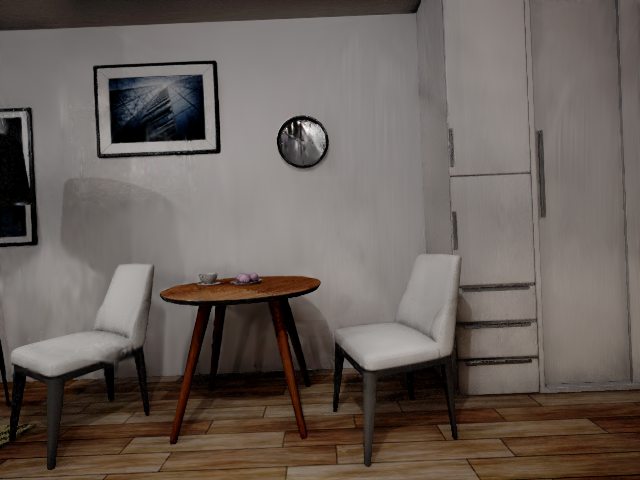}
  \includegraphics[width=\wratio\textwidth,height=\imgheight]{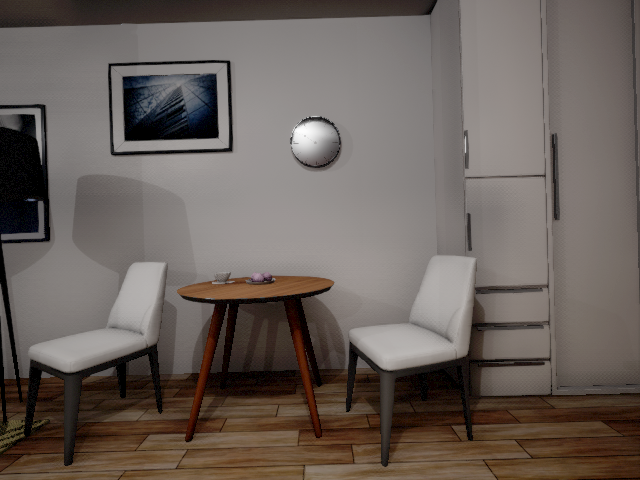}  

  \caption{\textbf{Qualitative Results.} Comparison across different long-term scene changes. (A) A black office chair appears in the scene; (B) the toy house and chair are moved, the picture is moved from the table to the shelf; (C) the cutlery on the table is replaced, the painting and the right chair are moved. \ours is the only method that captures the scene evolution and preserves high rendering quality.}
  \label{fig:qualitative_results}
\end{figure*}
\section{Experiments}\label{sec:experiment}

\boldparagraph{Datasets.} 
We test our method on the Flat~\cite{schmid2022panoptic} dataset, which consists of two RGB-D sequences captured in a synthetic environment with significant changes occurring between.
We further evaluate on the Aria~\cite{pan2023ariadigitaltwinnew} dataset to assess performance on real-world data. We select two recordings from two rooms each that have undergone long-term changes. Finally, the TUM-RGBD~\cite{sturm12iros} dataset is used to assess performance on three different static scenes following the protocol of~\cite{murai2024monogs}.

\boldparagraph{Evaluation Metrics.} To assess rendering quality, we compute PSNR,
SSIM~\citep{wang2004image} and LPIPS~\citep{zhang2018unreasonable}. Rendering metrics on all the datasets are evaluated by rendering full-resolution images along the ground-truth trajectory. We assess the depth error using the L1 norm in centimeters. In the tables, results are color-coded to indicate ranking: \fsttxt{best}, \sndtxt{second-best}, and \trdtxt{third-best}.

\boldparagraph{Baselines.} We compare \ours with state-of-the-art 3DGS online reconstruction systems MonoGS~\cite{murai2024monogs} and SplaTAM~\cite{keetha2024splatam}. We compare to DG-SLAM~\cite{lu20253dgs}, a recent dynamic 3DGS method, to assess the ability of dynamic 3DGS SLAM to handle evolving scenes. Wild-GS~\cite{zheng2025wildgs} cannot be run with ground-truth poses, as it relies on an off-the-shelf depth estimator that may be inconsistent with them. At the same time, it fails to track the camera poses on the Aria and Flat datasets. For this reason, we were not able to compare our method with it. 

\boldparagraph{Evolving Scene Evaluation Protocol.}
The goal of mapping evolving scenes is to render only the most up-to-date reconstruction without prior information on when the scene was changed. In addition, the system should be able to accurately render both the views it observed (input views) and unobserved frames (novel views). To achieve this, we merge RGB-D captures from every scene in the Aria and Flat datasets into a single continuous sequence to recreate this real-world behavior. For rendering evaluation, every 10th frame from each scene's last RGB-D sequence is held out for novel view synthesis testing. The remaining 90\% of frames are used to evaluate input view synthesis. To isolate the mapping performance, ground-truth poses are used for \ours and all baselines.
% ------------------------------------------------
\subsection{Evolving Scene Mapping Results}
We run \ours on the Flat and Aria datasets to evaluate rendering quality in evolving scenes, shown in~\cref{tab:rendering_flat_dataset,tab:rendering_aria_dataset}. The Flat dataset is designed to test mapping under scene changes and includes more substantial long-term dynamics. 
We note that all losses are computed over complete images, where changed areas typically occupy a smaller part.
Nonetheless, we observe notable differences in rendering performance on both datasets, where \ours is the only method that accurately adapts the reconstruction to even subtle scene changes without compromising rendering quality. 
This granularity is shown in qualitative results in~\cref{fig:qualitative_results}.
Beyond high rendering quality, \ours is able to accurately resolve changes also for challenging cases such as small cutlery on the table and flat paintings on the walls.
\begin{table}[t]
\centering
\resizebox{\linewidth}{!}{%
\begin{tabular}{lcccc}
\toprule
\textbf{Methods} & \textbf{PSNR [dB]} $\uparrow$ & \textbf{SSIM} $\uparrow$ & \textbf{LPIPS} $\downarrow$ & \textbf{Depth L1 [cm]} $\downarrow$ \\
\midrule
\textbf{SplaTAM}~\cite{keetha2024splatam} & \rd 15.88 / 12.69 & \rd 0.48 / 0.30 & \rd 0.55 / 0.70 & \rd 21.16 / 59.91 \\
\textbf{MonoGS}~\cite{murai2024monogs} & \nd 21.24 / 21.33 & \nd 0.77 / 0.77 & \nd 0.40 / 0.40 & \nd 30.95 / 29.91 \\
\textbf{DG-SLAM}~\cite{lu20253dgs} & 13.72 / 13.70 & 0.59 / 0.60 & 0.74 / 0.74 & 73.76 / 73.90 \\
\textbf{\ours} (Ours) & \fs 24.55 / 24.26 & \fs 0.93 / 0.93 & \fs 0.14 / 0.14 & \fs 6.9 / 7.9 \\
\bottomrule
\end{tabular}%
}
\caption{\textbf{Rendering performance on the Flat dataset.} Our method significantly outperforms other baselines thanks to dynamic scene adaptation and keyframe management system. Metrics are reported for input / novel views.}
\label{tab:rendering_flat_dataset}
\vspace{-2em}
\end{table}
\begin{table}[t]
\centering
\resizebox{\linewidth}{!}{%
\begin{tabular}{llcccc}
\toprule
\textbf{Methods} & \textbf{Scene} & \textbf{PSNR [dB]} $\uparrow$ & \textbf{SSIM} $\uparrow$ & \textbf{LPIPS} $\downarrow$ & \textbf{Depth L1 [cm]} $\downarrow$ \\
\midrule
\textbf{SplaTAM}~\cite{keetha2024splatam} 
& room0 & \rd 1.80 / \rd 16.75 & \rd 0.78 / \rd 0.45 & \rd 0.24 / \rd 0.44 & \rd 4.12 / \rd 17.79 \\
& room1 & \rd 22.94 / \rd 17.90 & \rd 0.81 / \rd 0.54 & \rd 0.22 / \rd 0.44 & \rd 2.84 / \rd 15.90 \\
& Avg   & \rd 22.37 / \rd 17.33 & \rd 0.80 / \rd 0.50 & \rd 0.23 / \rd 0.44 & \rd 3.48 / \rd 16.80 \\
\midrule
\textbf{MonoGS}~\cite{murai2024monogs} 
& room0 & \nd 25.28 / \nd 25.19 & \nd 0.78 / \nd 0.78 & \nd 0.28 / \nd 0.27 & \nd 5.15 / \nd 5.09 \\
& room1 & \nd 23.12 / \nd 23.02 & \nd 0.84 / \nd 0.84 & \nd 0.24 / \nd 0.24 & \nd 4.99 / \nd 5.01 \\
& Avg   & \nd 24.20 / \nd 24.11 & \nd 0.81 / \nd 0.81 & \nd 0.26 / \nd 0.26 & \nd 5.07 / \nd 5.05 \\
\midrule
\textbf{DG-SLAM}~\cite{lu20253dgs} 
& room0 & 15.78 / 15.63 & 0.58 / 0.58 & 0.76 / 0.76 & 67.60 / 69.05 \\
& room1 & 12.62 / 12.44 & 0.65 / 0.64 & 0.70 / 0.71 & 55.47 / 57.03 \\
& Avg   & 14.20 / 14.04 & 0.62 / 0.61 & 0.73 / 0.74 & 61.54 / 63.04 \\
\midrule
\textbf{\ours (Ours)} 
& room0 & \fs 31.54 / \fs 31.48 & \fs 0.95 / \fs 0.95 & \fs 0.14 / \fs 0.14 & \fs 0.69 / \fs 0.69 \\
& room1 & \fs 31.23 / \fs 31.97 & \fs 0.95 / \fs 0.95 & \fs 0.10 / \fs 0.10 & \fs 1.75 / \fs 1.80 \\
& Avg & \fs 31.39 / \fs 31.23 & \fs 0.95 / \fs 0.95 & \fs 0.12 / \fs 0.12 & \fs 1.22 / \fs 1.24 \\
\bottomrule
\end{tabular}%averages are still old!
}
\caption{\textbf{Rendering performance on the Aria dataset.} Our method significantly outperforms other baselines thanks to dynamic scene adaptaion and keyframe management system. Metrics are reported for input / novel views.}
\vspace{-1.5em}
\label{tab:rendering_aria_dataset}
\end{table}
\vspace{-1.5em}
\subsection{Ablation Studies}

\boldparagraph{Add and Remove operations are important for Dynamic Scene Adaptation}. We ablate the components of DSA in~\cref{tab:ablation_scene_adaptation}, showing that its presence significantly improves performance. While differences between individual DSA variants are less pronounced in final rendering quality, due to keyframe optimization correcting finer detail, combining both operations consistently yields the best performance.
\begin{table}[t]
\centering
\resizebox{\linewidth}{!}{%
\begin{tabular}{lcccc}
\toprule
\textbf{Method} & \textbf{PSNR [dB]} $\uparrow$ & \textbf{SSIM} $\uparrow$ & \textbf{LPIPS} $\downarrow$ & \textbf{Depth L1 [cm]} $\downarrow$ \\
\midrule
No DSA                  & 21.25 / 21.08 & 0.87 / 0.87 & 0.2 / 0.2 & 29.20 / 27.20 \\
Add                     & 23.34 / 23.07 & \textbf{0.92 / 0.91} & \textbf{0.15 / 0.15} & 11.00 / 11.20 \\
Remove                  & 24.13 / 23.89 & \textbf{0.93 / 0.93} & \textbf{0.14 / 0.14} & 9.50 / 8.20 \\
Add and Remove (Ours)   & \textbf{24.55 / 24.26} & \textbf{0.93 / 0.93} & \textbf{0.14 / 0.14} & \textbf{6.9 / 7.9} \\
\bottomrule
\end{tabular}%
}
\caption{\textbf{Dynamic Scene Adaptation (DSA) ablation on the Flat dataset.} Combining \textit{Add} and \textit{Remove} operations gives the best performance. Metrics are reported for input / novel views.}
\label{tab:ablation_scene_adaptation}
\vspace{-1em}
\end{table}

\boldparagraph{Our Keyframe Management system preserves useful information}. In~\cref{tab:ablation_keyframe_management}, we compare keeping all keyframes against ignoring stale keyframes and ignoring only stale regions. 
We observe that the naive approach to 3DGS mapping in evolving scenes by discarding conflicting keyframes can leave parts of the scene severely under-constrained, leading to the worst outcome.
Ignoring changes (no filtering) achieves clear background rendering (note that the background accounts for the majority of pixels and dominates the losses), but inevitably results in artifacts and inconsistencies for changed regions, as seen in \cref{fig:qualitative_results}.
Our proposed approach highlights the importance of retaining information about the background \emph{and} resolving conflicts in keyframes for 3DGS optimization.

\begin{table}
\centering
\resizebox{\linewidth}{!}{%
\begin{tabular}{lcccc}
\toprule
\textbf{Method} & \textbf{PSNR [dB]} $\uparrow$ & \textbf{SSIM} $\uparrow$ & \textbf{LPIPS} $\downarrow$ & \textbf{Depth L1 [cm]} $\downarrow$ \\
\midrule
No KF Filtering              & 23.57 / 23.31 & 0.92 / 0.92 & 0.15 / 0.14 & 6.70 / 7.20 \\
Full KF Filtering            & 13.72 / 13.34 & 0.58 / 0.57 & 0.56 / 0.55 & 764.40 / 763.30 \\
Partial KF Filtering (Ours)   & \textbf{24.55 / 24.26} & \textbf{0.93 / 0.93} & \textbf{0.14 / 0.14} & \textbf{6.9 / 7.9} \\
\bottomrule
\end{tabular}%
}
\caption{\textbf{Keyframe management ablation on the Flat dataset.} Retaining keyframes is essential to constrain the background, but conflicting regions must be accurately filtered to avoid artifacts.}
\label{tab:ablation_keyframe_management}
\vspace{-1.5em}
\end{table}

\boldparagraph{\ours can handle noisy poses}. 
We evaluate the camera poses on the Aria dataset with an external SLAM~\cite{murai2024monogs} system and use the estimated poses and keyframes instead of ground truth in~\cref{tab:supp_ablation_noisy_poses}. \ours is robust even when the pose estimation is not precise, exhibiting only a marginal performance drop under noisy pose conditions.

\begin{table}
\centering
\resizebox{\linewidth}{!}{%
\begin{tabular}{lcccc}
\toprule
\textbf{Camera Poses} & \textbf{PSNR [dB]} $\uparrow$ & \textbf{SSIM} $\uparrow$ & \textbf{LPIPS} $\downarrow$ & \textbf{Depth L1 [cm]} $\downarrow$ \\
\midrule
Ground-truth & \textbf{31.54 / 31.48} & \textbf{0.95 / 0.95} & \textbf{0.14 / 0.14} & \textbf{0.69 / 0.69} \\
Estimated    & 31.45 / \textbf{31.53} & \textbf{0.95 / 0.95} & \textbf{0.14 / 0.14} & 0.70 / 0.70 \\
\bottomrule
\end{tabular}%
}
\caption{\textbf{Effect of noisy poses on Aria room0.} \ours is robust to camera pose noise, showing no significant impact on performance when using estimated poses instead of ground truth. Camera poses and keyframes are estimated with an off-the-shelf method~\cite{murai2024monogs}.}
\label{tab:supp_ablation_noisy_poses}
\end{table}

\boldparagraph{\ours can reconstruct static scenes}. 
While our method is specifically designed to reconstruct evolving scenes, it should not lose the ability to reconstruct static scenes.
Results on the TUM-RGBD dataset shown in \cref{tab:ablation_static_scene} show that \ours performs on par with state-of-the-art systems.
This suggests that the introduced DSA and keyframe management are robust to noise and false positives in model updates and masking, which could deteriorate performance.
In the future, \ours could well be combined with tracking and registration methods to improve performance, however, this is currently beyond the scope of this work.
\begin{table}[t]
\centering
\resizebox{\linewidth}{!}{%
\begin{tabular}{lcccc}
\toprule
\textbf{Method} & \texttt{desk} & \texttt{xyz} & \texttt{office} & \textbf{Average} \\
\midrule
SplaTAM~\cite{keetha2024splatam} & \fs 20.92 & \fs 21.03 & \fs 21.61 & \fs 21.19 \\
MonoGS~\cite{murai2024monogs}    & \rd 17.41 & \rd 15.09 & \rd 19.93 & \rd 17.48 \\
\ours (Ours)                     & \nd 20.18 & \nd 21.01 & \nd 20.71 & \nd 20.63 \\
\bottomrule
\end{tabular}%
}
\caption{\textbf{Rendering performance on the TUM-RGBD dataset.} PSNR [dB]$\uparrow$ is reported for input views.}
\label{tab:ablation_static_scene}
\end{table}

\boldparagraph{\ours is an online mapping system}. 
While addressing scene changes requires additional computation time, our method performs on par with standard 3DGS reconstruction systems as shown in~\cref{tab:supp_ablation_runtime}. 
While the run time is not yet real-time, our approach is of an incremental nature with partial optimization performed on every new keyframe. 
We believe that future improvements in 3DGS optimization speed will translate well to \ours without incurring notable extra cost to handle evolving scenes.
Optionally, our approach lends itself to further final refinement. Both options are shown in~\cref{tab:ablation_global_refinement}.
While refinement can potentially improve the final rendering quality, we note that the model is already well converged after incremental processing.

\begin{table}
\centering
\resizebox{\linewidth}{!}{%
\begin{tabular}{lcccc}
\toprule
\textbf{Metric} & SplaTAM~\cite{keetha2024splatam} & MonoGS~\cite{murai2024monogs} & DG-SLAM~\cite{xu2024dgslam} & \ours (Ours) \\
\midrule
\textbf{FPS} $\uparrow$ & 0.13 & \fs 4.21 & \rd 0.25 & \nd 0.52 \\
\bottomrule
\end{tabular}%
}
\caption{\textbf{Runtime analysis on the Flat dataset.} FPS is calculated by dividing the mapping time by the total number of frames. All metrics are profiled on an NVIDIA RTX 3090 GPU.}
\label{tab:supp_ablation_runtime}
\vspace{-1em}
\end{table}
\begin{table}
\centering
\resizebox{\linewidth}{!}{%
\begin{tabular}{lcccc}
\toprule
\textbf{Method} & \textbf{PSNR [dB]} $\uparrow$ & \textbf{SSIM} $\uparrow$ & \textbf{LPIPS} $\downarrow$ & \textbf{Depth L1 [cm]} $\downarrow$ \\
\midrule
No Refinement      & 23.70 / 23.46 & 0.92 / 0.92 & 0.16 / 0.16 & \textbf{7.2 / 6.6} \\
With Refinement    & \textbf{24.55 / 24.26} & \textbf{0.93 / 0.93} & \textbf{0.14 / 0.14} & 7.9 / 6.6 \\
\bottomrule
\end{tabular}%
}
\caption{\textbf{Final refinement ablation on the Flat dataset.} While refinement can further improve rendering quality, \ours already converges well during incremental processing.}
\label{tab:ablation_global_refinement}
\vspace{-1.5em}
\end{table}

\boldparagraph{\ours is robust}.
Since the main focus of our work is on the architectural challenges when handling evolving scenes in 3DGS and not image-based change detection, we opted for a comparably simple threshold-based approach. 
While \ours is readily extendable with more complex methods such as~\cite{lu20253dgs}, we already find good performance and that \ours is robust w.r.t. critical hyperparameters.
First, most of our hyperparameters have an interpretable meaning, making them intuitive and requiring minimal effort to adjust. 
For instance, to determine $\epsilon_{\text{depth}}, \epsilon_{\text{color}}$, we inspected the rendering error in color and depth caused by adding a small object (e.g., a cup on the table) to the scene. 
For keyframe selection, we adopted standard rotation and translation thresholds~\cite{keetha2024splatam,yugay2023gaussianslam}. 
We experimentally show \ours to be robust w.r.t. hyperparameters by varying key parameters $\epsilon_{\text{depth}}, \epsilon_{\text{color}}, \epsilon_{\text{opacity}}$ with steps from the set of $\{-20\%, -10\%, 10\%, 20\%\}$ and report the mean and standard deviation of PSNR and Depth L1 error on the Flat dataset in~\cref{tab:hyperparameters}. 
The minimal variation in scores confirms that our method does not rely on sensitive parameter tuning.
\vspace{-1.5em}
\begin{table}
\centering
\resizebox{\linewidth}{!}{%
\begin{tabular}{lcccc}
\toprule
\textbf{Hyperparameter} & \textbf{PSNR} & \textbf{PSNR} & \textbf{Depth L1 [cm]} & \textbf{Depth L1 [cm]} \\
 & \textbf{Mean (↑)} & \textbf{Std. Dev. (↓)} & \textbf{Mean (↓)} & \textbf{Std. Dev. (↓)} \\
\midrule
$\epsilon_{\text{depth}}$ & 31.52 & 0.21 & 0.68 & 0.03\\
$\epsilon_{\text{color}}$ & 31.58 & 0.14& 0.69 & 0.01 \\
$\epsilon_{\text{opacity}}$ & 31.56 & 0.11 & 0.68 & 0.01 \\
\bottomrule
\end{tabular}%
}
\caption{\textbf{Hyperparameters robustness on the Aria room0 dataset.} \ours is robust to hyperparameter variation.}
\label{tab:hyperparameters}
\vspace{-1.5em}
\end{table}
\vspace{-0.3em}

%% file: sec/6_limitations.tex
\section{Limitations}
\label{sec:limitations}
While \ours currently demonstrates robust rendering performance in evolving long-term dynamic scenes, several notable limitations exist.
First, it does not yet handle short-term dynamic objects, which would result in inefficient addition and removal. Extending the dynamic scene adaptation mechanism to consider both dynamics is an exciting future direction.
Second, we primarily study mapping with external pose tracking.
Although sparse odometry systems can reject many changes as outliers~\cite{campos2021orb}, change-aware tracking mechanisms are an exciting future direction.
Finally, while \ours builds a consistent model of the scene at the final time, it would be interesting to explore a complete 4D representation of the history of all changes.
\vspace{-0.5em}

%% file: sec/7_conclusion.tex
\section{Conclusions}

We presented \ours, the first online mapping system for evolving scenes with novel view synthesis capabilities. 
\ours utilizes novel dynamic scene adaptation operations to detect and correct conflicts in the incrementally built 3DGS model. Our keyframe management method furthermore appropriately ignores stale areas in keyframes, retaining useful information while correcting for changed regions, resulting in a well-conditioned 3DGS optimization process.
We thoroughly evaluate \ours on synthetic and real-world datasets, demonstrating 29.7\% improvement in PSNR, over $3\times$-improvement in L1 depth error, and artifact-free novel view synthesis in evolving scenes. The code and data are released open-source.
\vspace{2em}

\boldparagraph{Acknowledgements.} This work was supported by TomTom, the University of Amsterdam, Amazon, ARL DCIST program, and the allowance of Top Consortia for Knowledge and Innovation (TKIs) from the Netherlands Ministry of Economic Affairs and Climate Policy.